\documentclass[11pt]{article}
\usepackage{fullpage,here,natbib,psfig}

\title{Models of Co-occurrence}
\author{I. Dan Melamed \\ Dept. of Computer and Information
Science \\ University of Pennsylvania \\ Philadelphia, PA, 19104,
U.S.A. \\ {\tt melamed@unagi.cis.upenn.edu} \\ {\tt http://www.cis.upenn.edu/\~{
}melamed}}

\newcommand{\ignore}[1]{}

\date{}

\begin{document}

\maketitle

\begin{abstract}
A {\bf model of co-occurrence} in bitext is a boolean predicate that
indicates whether a given pair of word {\em tokens} co-occur in
corresponding regions of the bitext space.  Co-occurrence is a
precondition for the possibility that two tokens might be mutual
translations.  Models of co-occurrence are the glue that binds methods
for mapping bitext correspondence with methods for estimating
translation models into an integrated system for exploiting parallel
texts.  Different models of co-occurrence are possible, depending on
the kind of bitext map that is available, the language-specific
information that is available, and the assumptions made about the
nature of translational equivalence.  Although most statistical
translation models are based on models of co-occurrence, modeling
co-occurrence correctly is more difficult than it may at first appear.
\end{abstract}

\section{Introduction}

Most methods for estimating translation models from parallel texts
(bitexts) start with the following intuition: Words that are
translations of each other are more likely to appear in corresponding
bitext regions than other pairs of words.  The intuition is simple,
but its correct exploitation turns out to be rather subtle.  Most of
the literature on translation model estimation presumes that
corresponding regions of the input bitexts are represented by neatly
aligned segments.  As discovered by
\citet{charalign}, most of the bitexts available today are not
easy to align.  Moreover, imposing an alignment relation on such
bitexts is inefficient, because alignments cannot capture crossing
correspondences among text segments.

\citet{simr} proposed methods for producing general bitext
maps for arbitrary bitexts.  The present report shows how to use
bitext maps and other information to construct a model of
co-occurrence.  A {\bf model of co-occurrence} is a boolean predicate,
which indicates whether a given pair of word {\em tokens} co-occur in
corresponding regions of the bitext space.  Co-occurrence is a
precondition for the possibility that two tokens might be mutual
translations.  Models of co-occurrence are the glue that binds methods
for mapping bitext correspondence with methods for estimating
translation models into an integrated system for exploiting parallel
texts.  When the model of co-occurrence is modularized away from the
translation model, it also becomes easier to study translation model
estimation methods {\em per se}.

Different models of co-occurrence are possible, depending on the kind
of bitext map that is available, the language-specific information
that is available, and the assumptions made about the nature of
translational equivalence.  The following three sections explore these
three variables.

\section{Relevant Regions of the Bitext Space}
\label{coocregions}

By definition of ``mutual translations,'' corresponding regions of a
text and its translation will contain word token pairs that are mutual
translations.  Therefore, a general representation of bitext
correspondence is the natural concept on which to build a model of
where mutual translations co-occur.  The most general representation
of bitext correspondence is a bitext map \citep{simr}.  Token pairs
whose co-ordinates are part of the true bitext map (TBM) are mutual
translations, by definition of the TBM.  The likelihood that two
tokens are mutual translations is inversely correlated with the
distance between the tokens' co-ordinate in the bitext space and the
interpolated TBM.

\begin{figure}[H]
\centerline{\psfig{figure=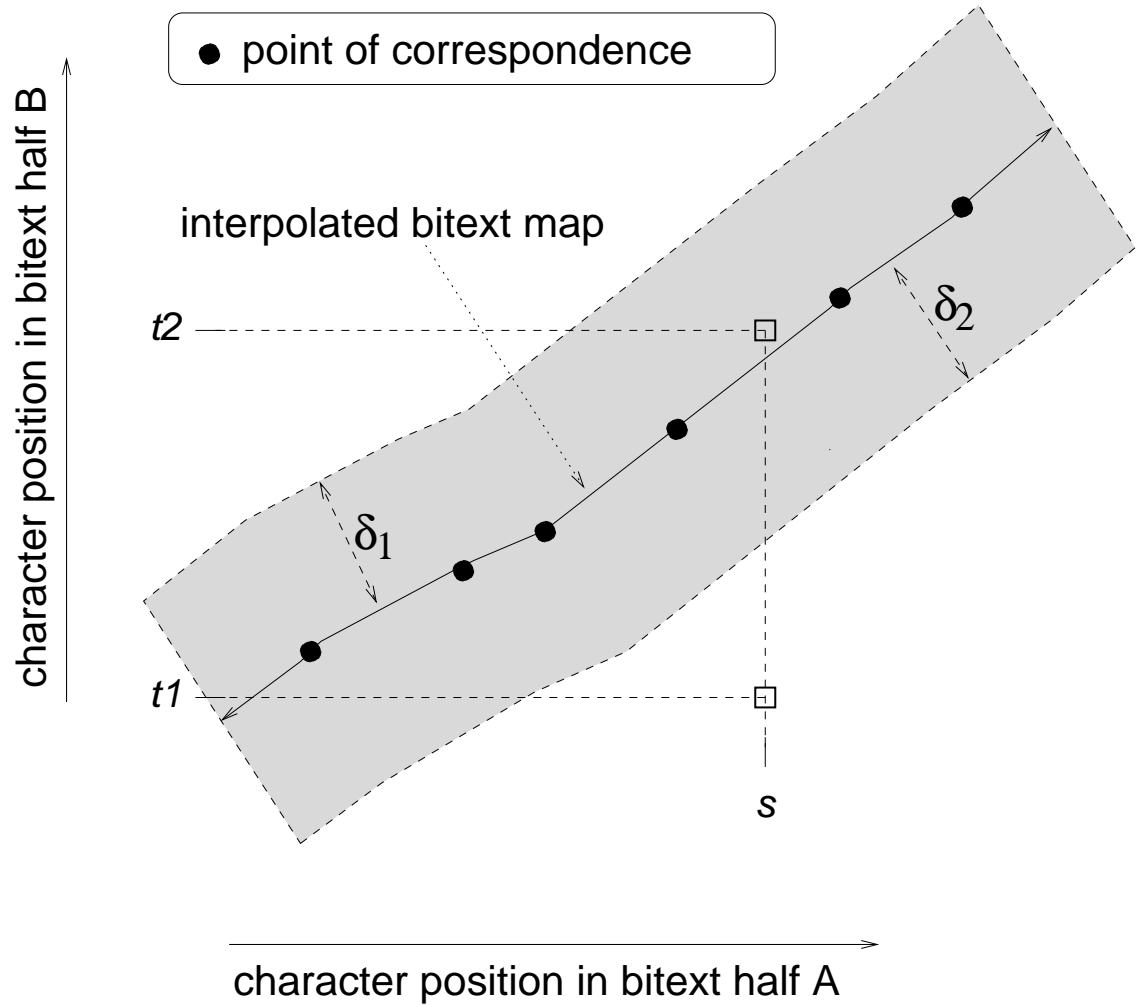,width=5.93in}}
\caption[{\em Distance-based model of co-occurrence}]{{\em
Distance-based model of co-occurrence.  Word token pairs whose
co-ordinates lie in the shaded region count as co-occurrences.  Thus,
$(s, t2)$ co-occur, but $(s, t1)$ do not.}\label{dcooc}}
\end{figure}

It may be possible to develop translation model estimation methods
that take into account a probabilistic model of co-occurrence.
However, all the models in the literature are based on a boolean
co-occurrence model --- they want to know either that two tokens
co-occur or that they do not.  A boolean co-occurrence predicate can
be defined by setting a threshold $\delta$ on the distance from the
interpolated bitext map.  Any token pair whose co-ordinate is closer
than $\delta$ to the bitext map would be considered to co-occur by
this predicate.  The optimal value of $\delta$ varies with the
language pair, the bitext genre and the application.
Figure~\ref{dcooc} illustrates what I will call the {\bf
distance-based model of co-occurrence}.  \citet{wordalign} were the
first to use a distance-based model of co-occurrence, although they
measured the distance in words rather than in characters.

\begin{figure}[H]
\centerline{\psfig{figure=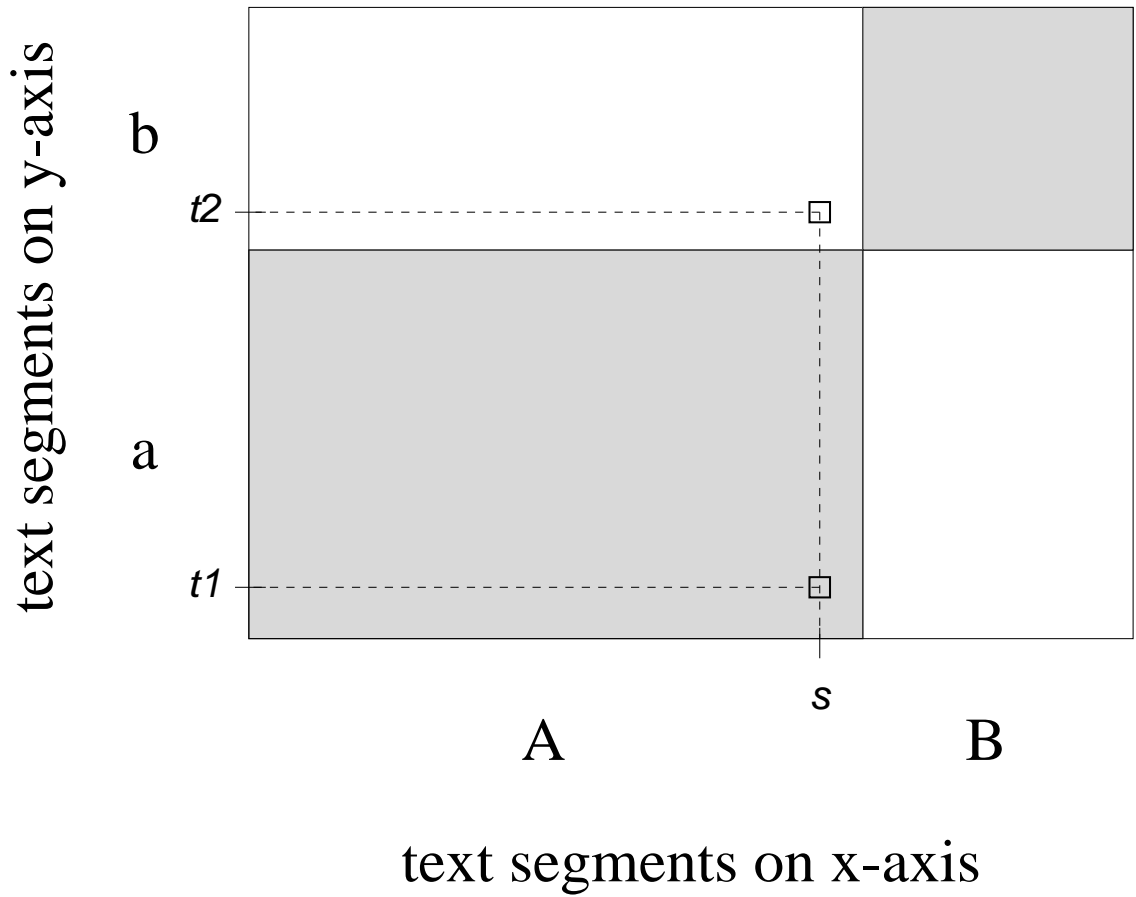,width=5.93in}}
\caption[{\em Boundary-based model of co-occurrence}]{{\em
Boundary-based model of co-occurrence.  Word token pairs whose
co-ordinates lie in shaded regions count as co-occurrences.  In
contrast with Figure~\ref{dcooc}, $(s, t1)$ co-occur, but $(s, t2)$ do
not.}\label{scooc}}
\end{figure}

General bitext mapping algorithms are a recent invention.  So far,
most researchers interested in co-occurrence of mutual translations
have relied on bitexts where sentence boundaries (or other text unit
boundaries) were easy to find \citep[\protect{\em e.g.}\
][]{wordcorr,kumano,funb,mel95}.  Aligned text segments suggest a {\bf
boundary-based model of co-occurrence}, illustrated in
Figure~\ref{scooc}.

\begin{figure}[H]
\centerline{\psfig{figure=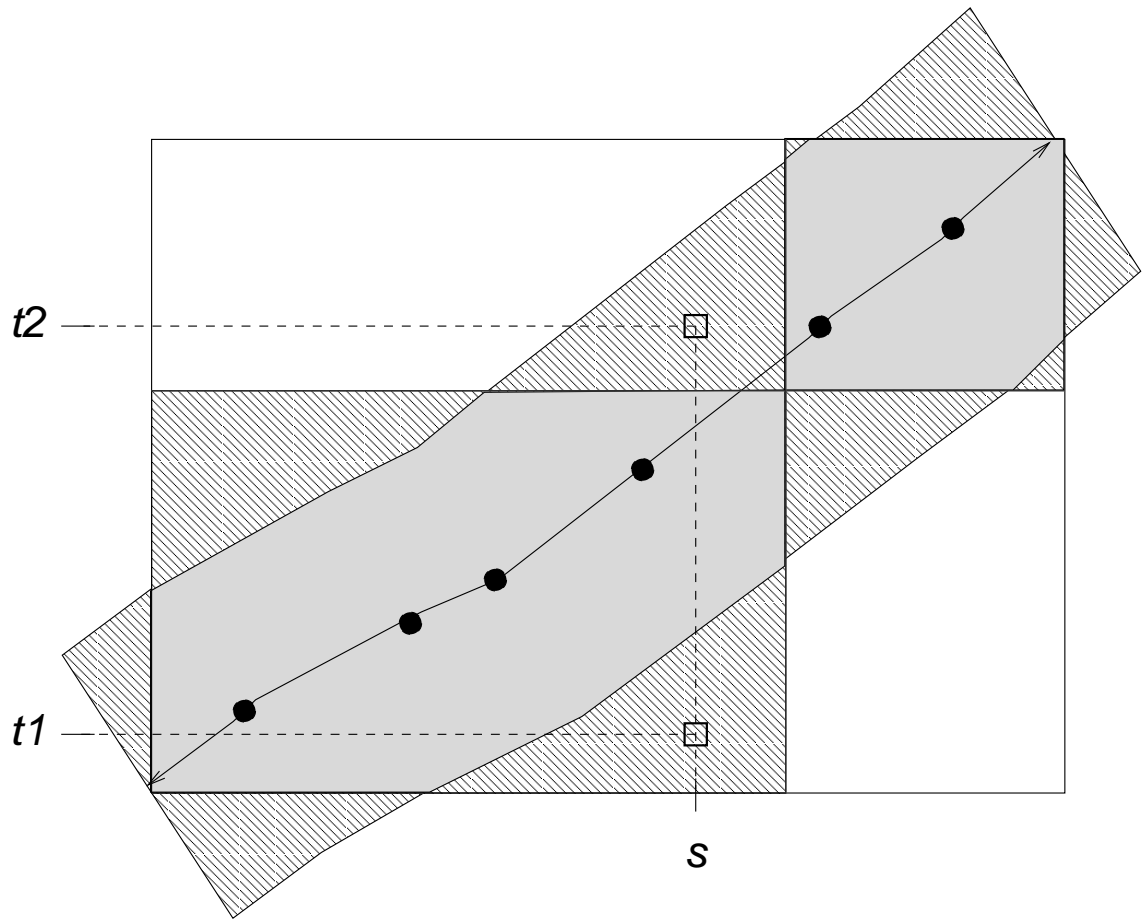,width=5.93in}}
\caption[{\em Combined model of co-occurrence}]{{\em Combined model of
co-occurrence.  Word token pairs whose co-ordinates lie in shaded
regions count as co-occurrences.  In contrast with Figures~\ref{dcooc}
and~\ref{scooc}, neither $(s, t1)$ nor $(s, t2)$ co-occur.  Striped
regions indicate eliminated sources of noise.}\label{bcooc}}
\end{figure}

For bitexts involving languages with similar word order, a more
accurate {\bf combined model of co-occurrence} can be built using both
segment boundary information and the map-distance threshold.  As shown
in Figure~\ref{bcooc}, each of these constraints eliminates the noise
from a characteristic region of the bitext space.

\clearpage

\section{Co-occurrence Counting Methods}
\label{countmethod}

Both the boundary-based and distance-based constraints restrict the
region of the bitext space where tokens may be considered to co-occur.
Yet, these constraints do not answer
the question of how to count co-occurrences within the restricted
regions.  It is somewhat surprising that this is a question at all,
and most authors ignore it.  However, when authors specify their
algorithms in sufficient detail to answer this question, the most
common answer \citep[given, \protect{\em e.g.},
by][]{ibm,wordalign,kupiec,mel95} turns out to be unsound.
The problem is easiest to illustrate under the boundary-based model of
co-occurrence.  Given two aligned text segments, the naive way to
count co-occurrences is
\begin{equation}
\label{naivecooc}
cooc({\bf u,v}) = e({\bf u}) \cdot f({\bf v})
\end{equation}
where $e({\bf u})$ and $f({\bf v})$ are the frequencies of occurrence
of ${\bf u}$ and ${\bf v}$ in their respective segments.  For many
{\bf u} and {\bf v}, $e({\bf u})$ and $f({\bf v})$ are either 0 or 1,
and Equation~\ref{naivecooc} returns 1 just in case both words occur.
The problem arises when $e({\bf u}) > 1$ and $f({\bf v}) > 1$.  For
example, if $e({\bf u}) = f({\bf v}) = 3$, then according to
Equation~\ref{naivecooc}, $cooc({\bf u,v}) = 9$!  If the two aligned
segments are really translations of each other, then it is most likely
that each of the occurrences of ${\bf u}$ is a translation of just one
of the occurrences of ${\bf v}$.  Although it may not be known which
of the 3 $v$'s each $u$ corresponds to, the number of times that ${\bf
u}$ and ${\bf v}$ co-occur as possible translations of each other in
that segment pair must be 3.

There are various ways to arrive at $cooc({\bf u,v}) = 3$.  Two of the
simplest ways are
\begin{equation}
\label{mincount}
cooc({\bf u,v}) = \min[e({\bf u}), f({\bf v})]
\end{equation}
and
\begin{equation}
\label{maxcount}
cooc({\bf u,v}) = \max[e({\bf u}), f({\bf v})] .
\end{equation}
Equation~\ref{mincount} is based on the simplifying assumption that
each word is translated to at most one other word.
Equation~\ref{maxcount} is based on the simplifying assumption that
each word is translated to at least one other word.  Either
simplifying assumption results in more plausible co-occurrence counts
than the naive method in Equation~\ref{naivecooc}.

Counting co-occurrences is more difficult under a distance-based
co-occurrence model, because there are no aligned segments and
consequently no useful definition for $e()$ and $f()$.  Furthermore,
under a distance-based co-occurrence model, the co-occurrence relation
is not transitive.  {\em E.g.}, it is possible that $s_1$ co-occurs
with $t_1$, $t_1$ co-occurs with $s_2$, $s_2$ co-occurs with $t_2$,
but $s_1$ does not co-occur with $t_2$.  The correct counting method
becomes clearer if the problem is recast in graph-theoretic terms.
Let the words in each half of the bitext represent the vertices on one
side of a bipartite graph.  Let there be edges between each pair of
words whose co-ordinates are closer than $\delta$ to the bitext map.
Now, under the ``at most one'' assumption of Equation~\ref{mincount},
each co-occurrence is represented by an edge in the graph's maximum
matching
\footnote{A {\bf maximum matching} is a subgraph that solves the
cardinality matching problem \citep[pp. 469-470]{netflow}.}.  Under
the ``at least one'' assumption of Equation~\ref{maxcount}, each
co-occurrence is represented by an edge in the graph's smallest vertex
cover.  Maximum matching can be computed in polynomial time for any
graph \citep{netflow}.  Vertex cover can be solved in polynomial time
for bipartite graphs\footnote{The algorithm is folklore, but
\citet{tandy} describe relevant methods.}.  It is of no importance that
maximum matchings and minimum vertex covers may be non-unique --- by
definition, all solutions have the same number of edges, and this
number is the correct co-occurrence count.

\section{Language-Specific Filters}
\label{langfilt}

Co-occurrence is a universal precondition for translational
equivalence among word tokens in bitexts.  Other preconditions may be
imposed if certain language-specific resources are available
\citep{mel95}.  For example, parts of speech tend to be preserved in
translation \citep{papag}.  If part-of-speech taggers are available
for both languages in a bitext, and if cases where one part of speech
is translated to another are not important for the intended
application, then we can rule out the possibility of translational
equivalence for all token pairs involving different parts of speech.
A more obvious source of language-specific information is a
machine-readable bilingual dictionary (MRBD).  If token $a$ in one
half of the bitext is found to co-occur with token $b$ in the other
half, and $(a,b)$ is an entry in the MRBD, then it is highly likely
that the tokens $a$ and $b$ are indeed mutual translations.  In this
case, there is no point considering the co-occurrence of $a$ or $b$
with any other token.  Similarly exclusive candidacy can be
granted to cognate token pairs \citep{simard}.

Most published translation models treat co-occurrence counts as counts
of potential link tokens \citep{mythesis}.  More accurate models may
result if the co-occurrence counts are biased with language-specific
knowledge.  Without loss of generality, whenever translation models
refer to co-occurrence counts, they can refer to co-occurrence counts
that have been filtered using whatever language-specific resources
happen to be available.  It does not matter if there are dependencies
among the different knowledge sources, as long as each is used as a
simple filter on the co-occurrence relation \citep{mel95}.

\section{Conclusion}

In this short report, I have investigated methods for modeling word
token co-occurrence in parallel texts (bitexts).  Models of
co-occurrence are a precursor to all the most accurate translation
models in the literature.  So far, most researchers have relied on
only a restricted form of co-occurrence, based on a restricted kind of
bitext map, applicable to only a limited class of bitexts.  A more
general co-occurrence model can be based on any bitext map, and thus
on any bitext.

The correct method for counting the number of times that two words
co-occur turns out to be rather subtle, especially for more general
co-occurrence models.  As noted in Section~\ref{countmethod}, many
published translation models have been based on flawed models of
co-occurrence.  This report has exposed the flaw and has shown how
to fix it.


\begin{thebibliography}{999}

\bibitem[\protect\citeauthoryear{Ahuja~{\em et~al.}}{1993}]{netflow}
        R. K. Ahuja, T. L. Magnati \& J. B. Orlin. (1993) {\em Network
        Flows: Theory, Algorithms, and Applications}. Prenice Hall,
        Englewood Cliffs, NJ.


\bibitem[\protect\citeauthoryear{Brown~{\em et~al.}}{1993}]{ibm} 
P. F. Brown, S. A. Della Pietra, V. J. Della Pietra, \& R. L. Mercer.
\newblock (1993)
\newblock ``The Mathematics of Statistical Machine
          Translation: Parameter Estimation,''
\newblock {\em Computational Linguistics 19}(2).


\bibitem[\protect\citeauthoryear{Church}{1993}]{charalign} 
K. W. Church.
\newblock (1993)
\newblock ``Char\_align: A Program
        for Aligning Parallel Texts at the Character Level,''
\newblock {\em
        Proceedings of the 31st Annual Meeting of the Association for
        Computational Linguistics}. Columbus, OH.

\bibitem[\protect\citeauthoryear{Dagan~{\em et~al.}}{1993}]{wordalign} 
I. Dagan, K. Church, \& W. Gale.
\newblock (1993)
\newblock ``Robust Word Alignment for Machine Aided Translation,''
\newblock {\em
        Proceedings of the Workshop on Very Large Corpora: Academic
        and Industrial Perspectives}. Columbus, OH.


\bibitem[\protect\citeauthoryear{Fung}{1995}]{funb}
        P. Fung. (1995) ``A Pattern Matching Method for Finding Noun
        and Proper Noun Translations from Noisy Parallel Corpora,''
        {\em Proceedings of the 33rd Annual Meeting of the Association
        for Computational Linguistics}. Boston, MA.

\bibitem[\protect\citeauthoryear{Gale~\&~Church}{1991}]{wordcorr} 
W. Gale \& K. W. Church.
\newblock (1991)
\newblock ``Identifying Word Correspondences in Parallel Texts,''
\newblock {\em Proceedings of the DARPA SNL Workshop}.
\newblock Asilomar, CA.

\bibitem[\protect\citeauthoryear{Kumano~\&~Hirakawa}{1994}]{kumano}
        A. Kumano \& H. Hirakawa. (1994) ``Building an MT Dictionary
        from Parallel Texts Based on Linguistic and Statistical
        Information,'' {\em Proceedings of the 15th International
        Conference on Computational Linguistics}.  Kyoto, Japan.

\bibitem[\protect\citeauthoryear{Kupiec}{1993}]{kupiec}
        J. Kupiec. (1993) ``An Algorithm for Finding Noun Phrase
        Correspondences in Bilingual Corpora,'' {\em Proceedings of
        the 31st Annual Meeting of the Association for Computational
        Linguistics}. Columbus, OH.

\bibitem[\protect\citeauthoryear{Melamed}{1995}]{mel95}
 I.~D. Melamed.
\newblock (1995)
\newblock ``Automatic evaluation and uniform filter cascades for inducing
  $N$-best translation lexicons,''
\newblock {\em Proceedings of the Third Workshop on Very Large Corpora}.
  Cambridge, Massachusetts.

\bibitem[\protect\citeauthoryear{Melamed}{1996}]{simr}
        I. D. Melamed. (1996) ``A Geometric Approach to Mapping Bitext
        Correspondence,'' {\em Proceedings of the First Conference on
        Empirical Methods in Natural Language Processing}.
        Philadelphia, PA.


\bibitem[\protect\citeauthoryear{Melamed}{1998}]{mythesis}
        I. D. Melamed. (1998) {\em Empirical Methods for Exploiting
        Parallel Texts}. Ph.D. dissertation, University of Pennsylvania,
        Philadelphia, PA.

\bibitem[\protect\citeauthoryear{Papageorgiou~{\em et~al.}}{1994}]
        {papag} H. Papageorgiou, L. Cranias \&
        S. Piperidis. (1994) ``Automatic Alignment in Parallel
        Corpora,'' {\em Proceedings of the 32nd Annual Meeting of the
        Association for Computational Linguistics (Student Session)}.
        Las Cruces, NM.

\bibitem[\protect\citeauthoryear{Phillips~\&~Warnow}{1996}]{tandy}
        C. Phillips \& T. J. Warnow. (1996) ``The Asymmetric median
        tree --- A New Model for Building Consensus Trees,'' {\em
        Discrete Applied Mathematics 71(1-3)}, pp.\ 331-335.


\bibitem[\protect\citeauthoryear{Simard~{\em et~al.}}{1992}]{simard} 
M. Simard, G. F. Foster \& P. Isabelle.
\newblock (1992)
\newblock ``Using Cognates to Align Sentences in Bilingual
        Corpora,''
\newblock  {\em Proceedings of the Fourth International
        Conference on Theoretical and Methodological Issues in Machine
        Translation}. Montreal, Canada.


\end{thebibliography}
\end{document}